# EFFICIENT LIKELIHOOD BAYESIAN CONSTRAINED LOCAL MODEL

*Hailiang Li*, Kin-Man Lam*, Man-Yau Chiu, Kangheng Wu, Zhibin Lei*
*Department of Electronic and Information Engineering, The Hong Kong Polytechnic University
Hong Kong Applied Science and Technology Research Institute Company Limited, Hong Kong, China
harley.li@connect.polyu.hk,{harleyli, edmondchiu, khwu, lei}@astri.org, enkmlam@polyu.edu.hk

## ABSTRACT

The constrained local model (CLM) proposes a paradigm that the locations of a set of local landmark detectors are constrained to lie in a subspace, spanned by a shape point distribution model (PDM). Fitting the model to an object involves two steps. A response map, which represents the likelihood of the location of a landmark, is first computed for each landmark using local-texture detectors. Then, an optimal PDM is determined by jointly maximizing all the response maps simultaneously, with a global shape constraint. This global optimization can be considered as a Bayesian inference problem, where the posterior distribution of the shape parameters, as well as the pose parameters, can be inferred using maximum a posteriori (MAP). In this paper, we present a cascaded face-alignment approach, which employs random-forest regressors to estimate the positions of each landmark, as a likelihood term, efficiently in the CLM model. Interpretation from CLM framework, this algorithm is named as an efficient likelihood Bayesian constrained local model (elBCLM). Furthermore, in each stage of the regressors, the PDM non-rigid parameters of previous stage can work as shape clues for training each stage regressors. Experimental results on benchmarks show our approach achieve about 3 to 5 times speed-up compared with CLM models and improve around 10% on fitting quality compare with the same setting regression models.

## 1. INTRODUCTION

The main goal of face alignment is to locate the semantic structural facial landmarks, such as eye brows, eyes, nose, mouth and face contour accurately (see Fig. 1). This information from facial landmarks is crucial to understand and analyze face relative works, such as expression recognition [9], face recognition [10], and face hallucination [11].

These classic Newton's gradient descent methods, such as active appearance model (AAM) [4] and constrained local models (CLM) [17, 18] try to use facial appearance (texture) information, i.e., the pixel intensity patterns in the area around the landmarks, and the face shape patterns which refer to the patterns of the face shape defined by the landmark coordinates to register facial landmarks. The shape model in ASM [3], AAM [4] and CLM is normally described by finding the parameters of a statistical shape distribution model, i.e., the point distribution model (PDM). The PDM is a simple yet efficient, linear model where facial shapes are model as a linear combination of eigen-shapes around the mean shape, which can generalize to unseen facial shapes.

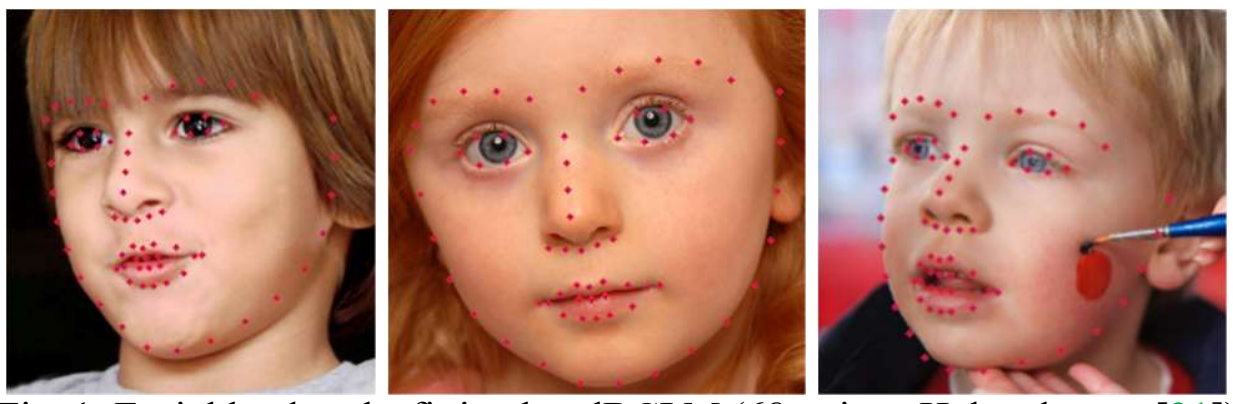

Fig. 1. Facial landmarks fitting by elBCLM (68 points, Helen dataset [21])

Recently a new family of face alignment algorithms has emerged [1, 2, 5, 8, 13], which directly learns regressors from image feature descriptors to the target shape gradually. These regression-based methods are gaining popularity, due to their excellent performance and high efficiency in face alignment task. The regression based methods do not explicitly learn any shape model, and they directly learn models mainly based on facial appearance or manually designed features and do prediction based on trained models.

In this paper, we propose a novel framework, which draws merits from these two categories. In our framework, the local response mapping part in CLM is replaced with random forest based regressors, which is able to widen the local area for calculating the local likelihood in the Bayesian framework and also significantly improve computation. This method can also be regarded as regression based methods with respect to a PDM shape constraint. From former opinion, we coined our novel algorithm as an efficient likelihood Bayesian constrained local model (elBCLM). With the PDM constraint, regression based method [5] with same setting can get 5% improvement on error management. Furthermore, to every stage for regressors, the PDM non-rigid parameters from pre-stage can work as shape clue on training each regression model, which achieve another 5% improvement.

The remainder of paper is organized as follows. In Section 2, previous relative arts are reviewed. In Section 3, CLM is given an overview; In Section 4, our proposed model, elBCLM, is described; Section 5 discusses an adaptive feature strategy to balance computation and efficiency in cascaded framework. Experiment results are presented in section 6 and a conclusion is given in Section 7.

## 2. PREVIOUS RELATIVE ARTS

Face alignment methods can be classified into two major categories, including the classic Newton's gradient descent based methods and the regression based methods. Typical Newton's gradient descent based methods include the active appearance model (AAM) [4], which try to learn the holistic model parameters by updating the Jacobian and Hessian matrix for later fitting stage, and the constrained local models (CLM) [6, 17, 19], which try to locate each landmark point independently through local appearance information by learning local patch response maps. The CLM model also embed the face shape model, i.e., the point distribution model (PDM) as the shape constraint. The PDM is a linear model with parameters to represent a set of shapes. It is used to estimate the likelihood of the points being in a model by given a set of feature points, which is important for model fitting, as it can act as a prior in the Bayesian framework.

Recently, regression based works [1, 2, 13,15] show significant better performances than the AAM and CLM frameworks. These models do not learn any shape model explicitly but only learn models for predicting landmarks directly from facial appearance feature patterns. The supervised decent method (SDM) [2] is one pioneer work on regression methods. SDM formulates the face alignment task as a general optimization problem, which is approximately solved by learning successive mapping functions from local appearance feature patterns, such as histogram of oriented gradients (HOG [19]) or scale-invariant feature transform (SIFT [7]) features, to the shape updates with linear regression models. In [1], the authors present a highly efficient local binary features (LBF), which derived from local pixel shape-indexed features, with linear regression framework achieve faster speed with comparable quality. Paper [5] proposes a local lightweight feature, namely intimacy definition feature (IDF) which obtains about two times the speed-up and more than 20% improvement in terms of alignment error measurement than LBF [1] method.

The regression based methods take advantage of shape information in a limited sense in a way that all the points are updated jointly, i.e., each point is with linear regression with features from all other points, so the shape pattern constraint is implicitly embedded in the model. While since there is no explicitly using with the shape prior, one major limitation of the current regression based methods is the ignorance or ineffective usage of the shape information.

## 3. CONSTRAINED LOCAL MODELS

### 3.1. The Shape Model -PDM model

The shape X of a point distribution model (PDM) is represented by the 2D vertex locations of a mesh, with a $2n$ dimensional vector: $\mathrm{X} = (x_1, y_1, .., x_n, y_n)^{\mathrm{T}}$. The traditional way of building a PDM requires a set of shape annotated images that are previously aligned in scale, rotation and translation by Procrustes analysis [16]. Applying a PCA to a set of aligned training examples, the shape can be expressed by the linear parametric model:

$$X_i = s\ \ (\bar{X}_i + \Phi_i \mathrm{q}) + t \qquad (1)$$

where $X_i$ denotes the 2D-location of the PDM's $i^{th}$ landmark and $\mathbf{p} = \{s, \mathbf{R}, \mathbf{t}, \mathbf{q}\}$ denotes the PDM parameters, which consist of a global scaling $s$, a rotation $\mathbf{R}$, a translation $\mathbf{t}$ and a set of non-rigid parameters $\mathbf{q}$. Here, $\bar{\boldsymbol{X}}_i$ denotes the mean location of the $i^{th}$ PDM landmark in the reference frame (i.e. $\bar{\boldsymbol{X}}_i = [\bar{x}_i; \bar{y}_i]$ for a 2D model) and $\Phi$ is the shape subspace matrix holding $n$ eigenvectors (retaining a user defined variance, e.g. 99%), so $\mathbf{q}$ can be assumed as a vector of shape parameters. From the probabilistic point of view, the non-rigid shape parameters $\mathbf{q}$ exhibit a Gaussian distribution leading to the following prior:

$$p(\mathbf{q}) \propto \mathcal{N}(\mathbf{q}; \mathbf{0}, \mathbf{\Lambda}); \ \ \mathbf{\Lambda} = \mathrm{diag}\{[\lambda_1; \lambda_2; \dots; \lambda_{\mathrm{m}}]\} \quad (2)$$

where $\lambda_i$ denotes the PCA eigenvalue of the $i^{th}$ mode of deformation. $\mathbf{\Lambda}$ is constructed from the training set, based on how much shape variation in the training is explained by the $i^{th}$ parameter, with $\lambda_i$ corresponding to the $\mathbf{q}_i$ parameter.

In this paper, we refer methods as constrained local models (CLM) as an algorithm to that utilizes an ensemble of local detectors with PMD shape model constraint. The CLM models have attracted some interest as it avoids many of drawbacks of holistic approaches. Based on this definition a CLM normally consists of two parts: a statistical shape model and patch experts (also called local detectors). Both the shape model and patch experts can be trained offline and then used for online landmark detection, which is achieved by fitting the CLM to a given image. The deformable model is controlled by parameters $\mathbf{p}$ and the instance of a model can be described by the locations of its feature points $X_i$ in an image $I$ as shown in Fig. 2.

### 3.2. CLM model and fitting process

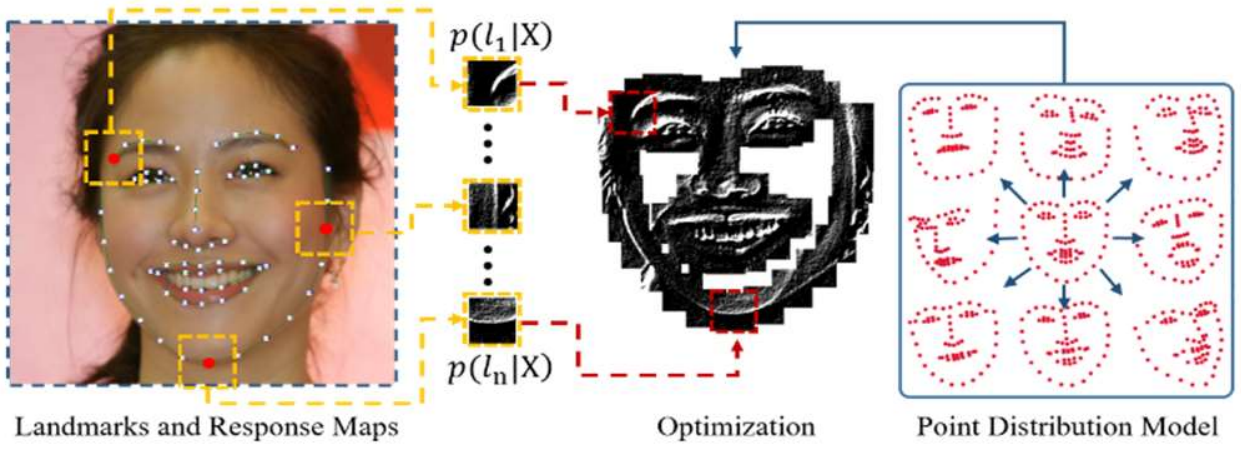

Fig. 2 Overview of CLM fitting process

CLM fitting is generally posed as search for the PDM parameters, $\mathbf{p}$, that jointly minimizes the misalignment error over all landmarks, regularized appropriately:

$$\mathcal{E}(\mathbf{p}) = R(\mathbf{p}) + \sum_{i=1}^{n} D_i(\boldsymbol{X}_i; I) \qquad (3)$$

where $R$ penalizes complex deformations (i.e. the regularization term) and $D_i$ denotes the measure of misalignment for the $i^{th}$ landmark at $\boldsymbol{X}_i$ in the image $I$ (i.e. the data term). The form of regularization is related to the assumed distribution of PDM parameters describing plausible object shapes, common examples of which include the Gaussian and Gaussian mixture model (GMM) estimates.

As shown in Fig. 2, the CLM fitting processing mainly contains two components:

(1) an exhaustive local search for feature locations to get the response maps:

$$\{p(l_i = aligned|I, x)\}_{i=1}^{n} \tag{4}$$

(2) an optimization strategy to maximize the responses of the PDM constrained landmarks.

Most innovations to the CLM model is the replacement of the distribution of probable landmark locations, obtained from each patch based local detector, i.e., to replace the response filters with simpler and more accurate ones. For the optimization step: Once response maps for each landmark have been found, by assuming conditional independence, optimization proceeds by maximizing:

$$p(\{l_i = aligned\}_{i=1}^{n}|p) = \prod_{i=1}^{n} p(l_i = aligned|x_i) \tag{5}$$

with respect to the PDM parameters $\mathbf{p}$, where $\boldsymbol{X}_i$ is parameterized as in Equation (1) and dependence on the image $I$ is dropped for succinctness. It should be noted that some forms of CLMs pose Equation (4) as minimizing the summation of local energy responses.

Where $l_i$ is a discrete random variable denoting whether the $i^{th}$ landmark is correctly aligned or not, $I$ is the image.

### 3.3. CLM in Bayesian formulation

The CLM objective in Eqn. 5 can be interpreted as maximizing the likelihood of the model parameters such that all of its landmarks are aligned with their corresponding locations on the object in an image. The specific form of the objective implicitly assumes conditional independence between detections for each landmark, the probabilistic interpretation of which takes the form:

$$p(\mathbf{p}|\{l_i = 1\}_{i=1}^{n}, I) \propto p(\mathbf{p}) \prod_{i=1}^{n} p(l_i = 1|x_i, I) \tag{6}$$

i.e.: $\ln\{p(\mathbf{p}|\{l_i = 1\}_{i=1}^{n}, I)\} \propto \ln\{p(\mathbf{p})\} +$

$$\sum_{i=1}^{n} \mathbf{ln}\{p(l_i = 1|x_i, I)\} \tag{7}$$

based on Equation (2), we can get:

$$\mathcal{E}(\mathbf{p}) = -\ln\{p(\mathbf{p})\} \tag{8}$$

$$D_i(\boldsymbol{X}_i; I) = -\mathbf{ln}\{p(l_i = 1|x_i, I)\} \tag{9}$$

When assuming a non-informative (uniform) prior over the PDM parameters, the formulation in (6) leads to a maximum likelihood (ML) estimate, otherwise it leads to a maximum *a-posterior* (MAP) estimate.

The method entails first finding the location within each response map for which the maximum was attained: $\mu = [\mu_1; \dots; \mu_n]$. The objective of the optimization procedure is then to minimize the weighted least squares difference between the PDM and the coordinates of the peak responses, regularized appropriately:

$$\mathcal{E}(\mathbf{p}) = ||\mathbf{q}||^{2}_{\Lambda^{-1}} + \sum_{i=1}^{n} w_i ||\boldsymbol{X}_i - x\boldsymbol{\mu}_i||^2 \tag{10}$$

where the weights $\{w_i\}_{i=1}^{n}$ reflect the confidence over peak response coordinates and are typically set to some function of the responses at $\{\boldsymbol{\mu}_i\}_{i=1}^{n}$ =1, making it more resistant towards such things as partial occlusion, where occluded landmarks will be more weakly weighted. Equation (10) is iteratively minimized by taking a first order Taylor expansion of the PDM's landmarks:

Equation (10) is iteratively minimized by taking a first order Taylor expansion of the PDM's landmarks:

$$\boldsymbol{X}_i \approx X_i^c + \mathbf{J}_i \boldsymbol{\Delta}\mathbf{p} \tag{11}$$

and solving for the parameter update:

$$\boldsymbol{\Delta}\mathbf{p} = -\mathbf{H}^{-1}\boldsymbol{X}_i(\boldsymbol{\Lambda}^{-1}\mathbf{p} + \sum_{i=1}^{n} w_i \mathbf{J}_i (X_i^c - \boldsymbol{\mu}_i)) \tag{12}$$

which is updated to current parameters: $\mathbf{p} \leftarrow \mathbf{p} + \boldsymbol{\Delta}\mathbf{p}$. Here, $\boldsymbol{\Lambda} = \boldsymbol{diag}\{[\mathbf{0}; \boldsymbol{\lambda}_1; \boldsymbol{\lambda}_2; \dots; \boldsymbol{\lambda}_m]\}$, $\mathbf{J} = [\mathbf{J}_1; \mathbf{J}_2; \dots; \mathbf{J}_n]$ is the PDM's Jacobian, $\boldsymbol{X}^c = [X_1^c; X_2^c; \dots; X_n^c]$ is the current shape estimate and

$$\mathbf{H} = \boldsymbol{\Lambda}^{-1} + \sum_{i=1}^{n} w_i \mathbf{J}_i^T \mathbf{J}_i \tag{13}$$

is the Gauss-Newton Hessian matrix. For the partial derivative Jacobian matrix, there are coordinates of n landmarks partial derivative over the PDM parameters. i.e., 4 global rigid parameters ($s$, $\theta$, $t_x$, $t_y$), where $\theta$ is the angel of rotation matrix $\boldsymbol{R}$ plus $m$ local non-rigid parameters $\mathbf{q}$, where $m$ is the number of the eigenvectors after PCA for the whole training sampled shapes.

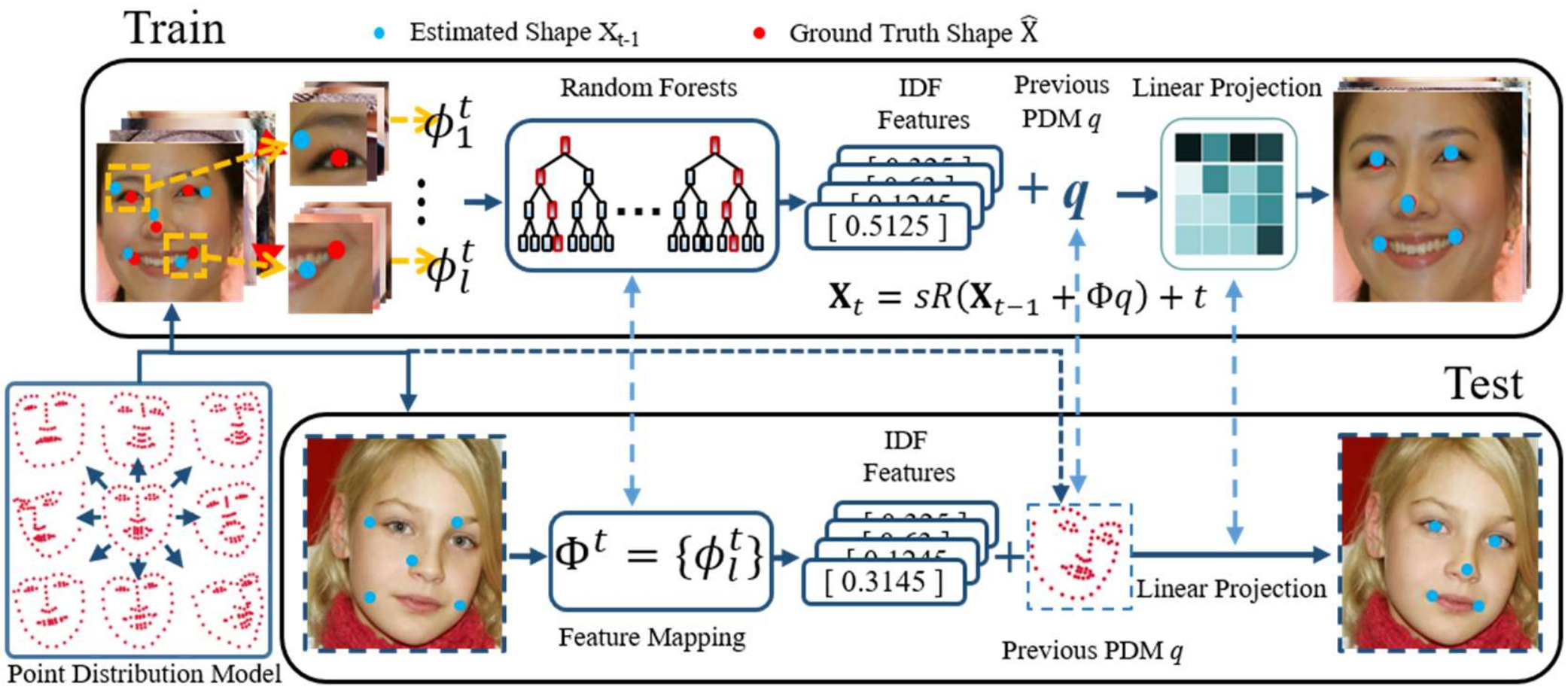

Fig. 3: An overview of the workflow for elBCM cascaded regression face alignment

## 4. PROPOSED MODEL

In recent years, random forests [14] have emerged as very useful classifiers for a large variety of computer vision tasks. This method is relatively simple and has many merits that make it particularly interesting for computer vision problems, more detail can reference to paper [5].

### 4.1. Random Forests based cascaded shape regression

Many face alignment methods work under a cascaded framework, where an ensemble of $N$ regressors operates in a stage-by-stage manner, which are referred to as stage regressors. This approach was first explored in [8]. At the testing stage, the input to a regressor $R_t$ at stage $t$ is a tuple $(I, X_{t-1})$ , where $I$ is an image and $X_{t-1}$ is the shape estimate from the previous stage (the initial shape $X_0$ is typically the mean shape of the training set). The regressor extracts features, i.e., they depend on the current shape estimate, or other features with respect to the current shape estimate, and regresses a vector of shape increment as follows:

$$X_t = X_{t-1} + R_t\big(\phi_t(I, X_{t-1})\big) \tag{14}$$

where $\phi_t(I, X_{t-1})$ can be referred to as the shape-indexed features which derived from shape-indexed features, such as LBF [1] and IDF [5]. The cascade progressively infers the shape in a coarse-to-fine manner – the early regressors handle large variations in shape, while the later ones ensure small refinements. After each stage, the shape estimate resembles the true shape closer and closer.

In our proposed algorithm, the feature mapping function $\phi_t(I, X_{t-1})$ generates local IDF features which derived from the shape-indexed feature or HOG features which extracted from the estimated landmark positions. There is an assumption, proved by intensive experimental results, that the shape increments have close correlation with the local features of the landmarks which define the face shape. Thus, given the features and the target shape increments $\{\Delta X_t = X_t - X_{t-1}\}$, a linear projection matrix $R_t$ can be learned. Most regression models [1, 2, 5, 8] share similar workflow.

### 4.2. Motivation and proposed method

If the weights are set to same and iteration works only one time, then equation (12) can be a simple version as:

$$\Delta \mathrm{p} = -\mathrm{H}^{-1}\mathrm{X}(\Lambda^{-1}\mathrm{p} + \mathrm{J}v) \tag{15}$$

Here $\boldsymbol{v}$ is the shape shift vector obtained by any response mapping algorithm. Recently researcher worked for getting different response mapping algorithm [18]. Obtaining shape shift vector $\boldsymbol{v}$ only once can break the hinder of local response mapping filters with other algorithm which can achieve the shift vector as setting $v = \Delta\mathrm{X}$, while the accuracy can be compensated by iterations from the cascaded framework:

$$\Delta \mathrm{p} = -\mathrm{H}^{-1}\mathrm{X}(\Lambda^{-1}\mathrm{p} + \mathrm{J}\Delta\mathrm{X}) \tag{16}$$

Equation (16) interprets the main idea of our proposed method, in which random forests based regressors are trained to obtain $\Delta$X efficiently, then update $\Delta$p for refining current fitted shape with PDM model constraint. Since response mapping filters mainly work in the way as convolutional filter, the drawback of convolutional filter significantly hinder CLM algorithm: if the size of patch based response mapping window set too small, the response mapping filter cannot cover bigger area to handle large posed faces; on the other hand, if set too big, convolutional computation may greatly reduce algorithm's efficiency. The PDM based prior term as in equation (2) according to the approximations taken, can be written as:

$$p(\mathbf{q}_k|\mathbf{q}_{k-1}) \propto \mathcal{N}\big(\mathbf{q}_k|\mu_q, \Sigma_q\big); \tag{17}$$

where $\mu_q = \mathbf{q}_{k-1}$ and $\Sigma_q = \boldsymbol{\Lambda}$ . This form of prior assumption can be largely improved recursively during the cascaded framework.

Inspired by this analysis, we propose a novel variant CLM algorithm by replacing response mapping filters with regressors to get shift vector $\Delta$X, and let refinement of shift vector $\Delta$X implicitly worked in the cascade framework. Interpretation on CLM opinion, random forest based regressors are used to replace the local response maps as an efficient likelihood calculation in the concept of Bayesian framework. From this opinion, we named our proposed novel algorithm as an efficient likelihood Bayesian constrained local model (elBCLM).

Our proposed idea, using regressors to replace local response maps computing has following advantages: (1) This method can circumvent hypothesis that each local detector is assumed conditional independent; (2) This method is able to avoid local optima which may be caused by typically local noise and ambiguities since small image patches often contain limited structure; (3) This method is capable of extending response map computation with more efficient methods. The whole workflow of elBCLM algorithm is described in Fig. 3, and the random forests based regression part can be referenced to similar works [1, 5].

### 4.3. PDM works as prior for regression features

The PDM is a linear model which parametrizes a class of shapes. It can also be used to estimate the likelihood of the points being in a model, given a set of feature points. This is important for model fitting, as it can act as a prior which works as a close-loop in the cascade framework.

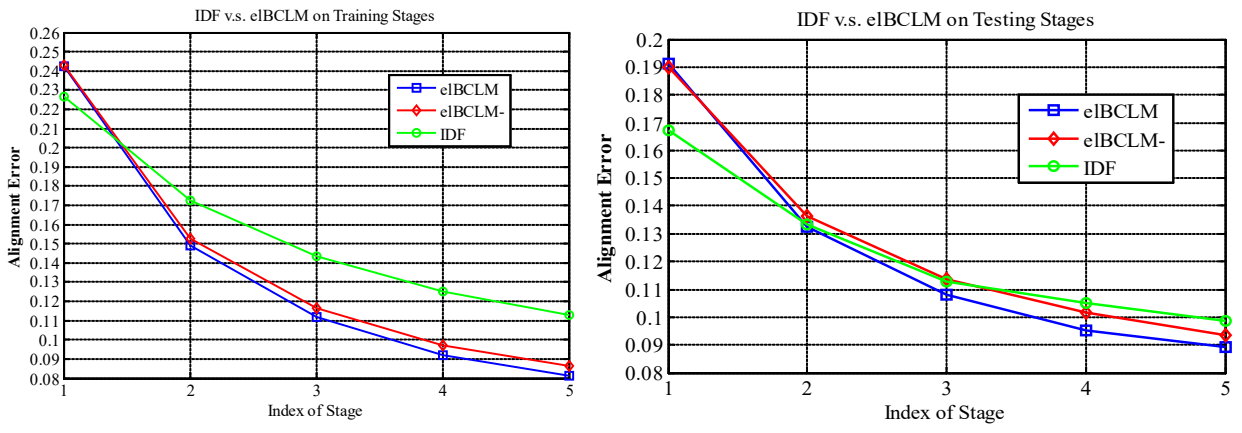

Fig. 4: regression based method IDF[5] compare with elBCLM-(only with PDM shape constraint) and elBCLM

The PDM $\mathbf{p} = \{s, \mathbf{R}, \mathbf{t}, \mathbf{q}\}$ can separate as two parts, global parameter: $\{s, \mathbf{R}, \mathbf{t}\}$ part, which consist of a global scaling $s$, a rotation $\mathbf{R}$, a translation $\mathbf{t}$ and the local part, a set of non-rigid parameters $\mathbf{q}$. As the local parameters $\mathbf{q}$ has more relation between the consecutive fitting shapes, so the previous local parameters $\mathbf{q}$ can work as guided shape feature for fitting current stage shape. In our experimental, for 68 facial landmarks, total dimension can 136 (=68*2) normally reduced to around 32 dimension by keeping 99% energy after PCA, so parameters $\mathbf{q}$ is a 32 vector. As $\mathbf{q}$ is added as regression features, which means the feature extraction part: $\phi_t(I, X_{t-1})$ in Eqn. (16) is refined.

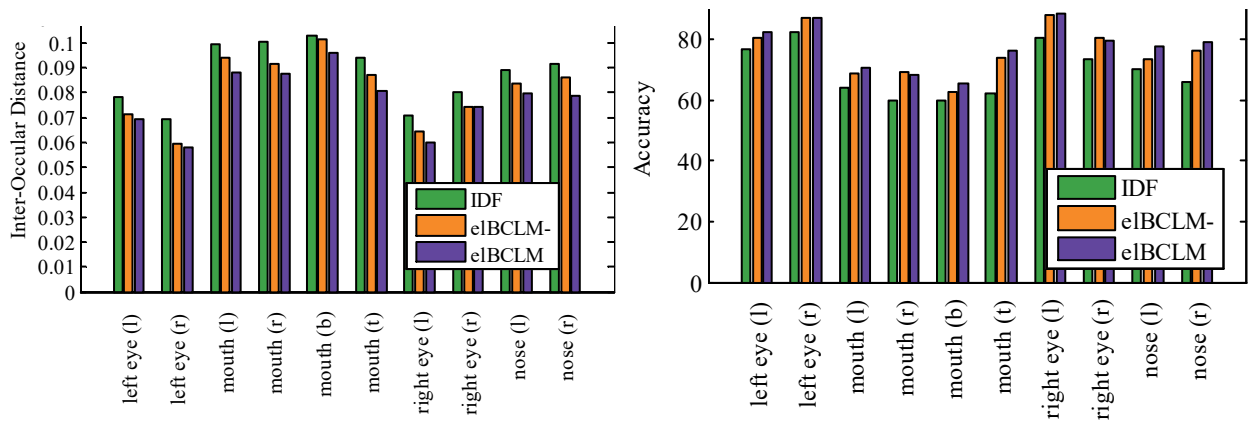

Fig. 5: regression based method IDF[5] compare with elBCLM-(only with PDM shape constraint) and elBCLM in 10 facial landmarks

As experimental results shown in Fig. 4, both on the training and testing stage, when regression algorithm pluses PDM shape constraint (as shown in the curve of elBCLM-), the alignment error can reduce more than 5% with same setting on the fitting stage; While once added the PDM model's local parameters $\mathbf{q}$ as features for training the linear regression models in each stage, the algorithm gets additional 5% gain on alignment error on the fitting stage.

The performance on accuracy and Inter-Occular distance criterion in Fig. 5 also show the same trend with three different configurations.

## 5 ADAPTIVE FEATURE

For cascade framework, there is some strategies to fine-tune the performance of the regression part. There are already some adaptive configurations during the process. Such as block window size adaptive, i.e., larger stage index, smaller window size to balance the computation [1, 2, 5]. Paper [19] proposes regression adaptive: to enhance the capability of handling large variation, for all stages, a more flexible mode of adaptive regression: global regression ⇒ part regression ⇒ local regression mechanism.

Studied from experimental results, we found simple and complex features both have relatively faster converge capability in the earlier stages, see in Fig. 6. While after earlier stages, the discrimination of the feature will not be sensitive, so balance speed and accuracy, there is a timing to shift to other more discriminative features in latter stages. Also as shown in Fig. 6, for the hybrid version, we configure 7 stages as first 4 stages with IDF features to get gain a faster speed and set latter 3 stages with HOG features to get higher accuracy by sacrificing little efficiency. This configuration gives out a relative more accurate and faster alignment algorithm compare some state-of-arts methods.

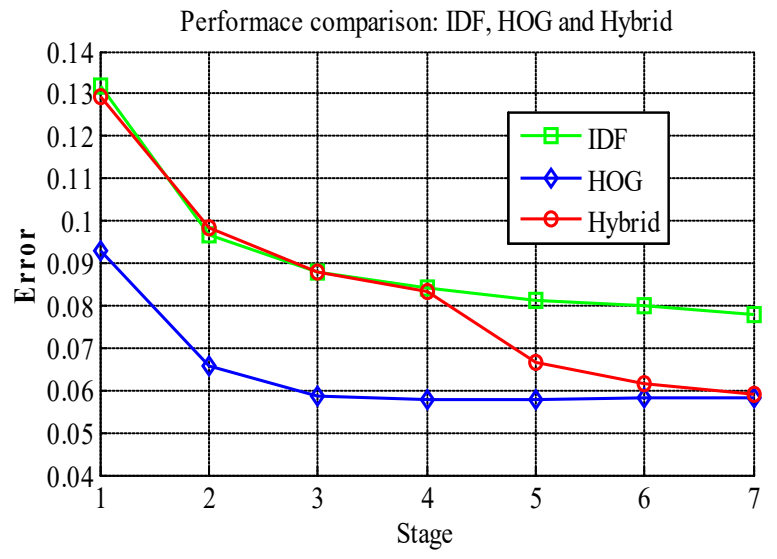

Fig. 6: Alignment comparison: IDF, HOG and Hybrid (Helen dataset [21])

## 6. WORKFLOW AND EXPERIMENTAL RESULTS

### 6.1. Algorithm workflow

The two stages of our proposed algorithm are described in Algorithm 1 and Algorithm 2 respectively.

**Algorithm 1: elBCLM Training Stage:**

**Input:** PDM $(X, \Phi)$ model, training data$(I_i, X_i, \bar{X}_i)$, for i=$1, \ldots, N$, where $I$ are face images, and $X$ are shapes; $N$ is the number of samples.
**Output:** regressors: $R = (R_1, \ldots, R_T)$, $T$: stage count.
**for** $t$=1 to $T$ **do**
  **for** all $i \in (1 \ldots N)$ **do**
    $\Delta X_t^i = X_t^i - \bar{X}_t^i$ ⇒ calculate $\Delta X_t^i$
    $f_t^i = \phi_t(I^i, S_{t-1}^i)$ ⇒ features + PMD's $\mathbf{q}$
  **end for**
  $R_t = \arg min_R \sum_i |R(f_t^i) - \Delta X_t^i|$
  **for** all $i \in (1 \ldots N)$ **do**
    $\bar{X}_t^i = \bar{X}_t^i + R(f_t^i)$ ⇒ update shape {Eqn. (14)}
    update $\Delta$p with $\Delta S_t^i$ ⇒ {Eqn. (16)}
    $\bar{X}_t^i \approx \bar{X}_t^i + \mathrm{J}_t \Delta \mathrm{p}$ ⇒ update shape w.r.t: Eqn. (10)
  **end for**
**end for**

**Algorithm 2: elBCLM Fitting Stage:**

**Input:** PDM $(X, \Phi)$ model, testing image $I$, initial (mean) shape $X^0$, trained regressors: $R = (R_1, \ldots, R_T)$
**Output:** Estimated pose $X^T$
**for** $t$=1 to $T$ **do**
  $f_t = \phi_t(I, X_{t-1})$ ⇒ features + PMD's $\mathbf{q}$
  $\Delta X = R_t(\phi_t(I, X_{t-1}))$ ⇒ apply regressor $R_t$
  $X_t = X_t + \Delta X$ ⇒ update shape {Eqn. (14)}
  update $\Delta$p with $\Delta X$ ⇒ {Eqn. (16)}
  $X_t \approx X_t + \mathrm{J}_t \Delta \mathrm{p}$ ⇒ update shape w.r.t: Eqn. (10)
**end for**

### 6.2. Experimental Results

Experiments on Helen dataset [21], elBCLM achieves with 5.88 alignment error, 150 FPS on 68 facial landmarks, which comparable to some state-of-art algorithms (see Table-1). Fig. 7 are some results of elBCLM algorithm compare with algorithm LBF [1] and CLM [18], which shows elBCLM can work more accurately in alignment quality.

| Method | Error (68 landmarks) |
|---|---|
| Zhu et. al [20] | 8.16* |
| DRMF [12] | 6.70* |
| RCPR [6] | 5.93* |
| Tadas et. al [18] | 6.75 |
| elBCLM | 5.88 |

Table-1: Alignment comparision; Reported from original with "*".

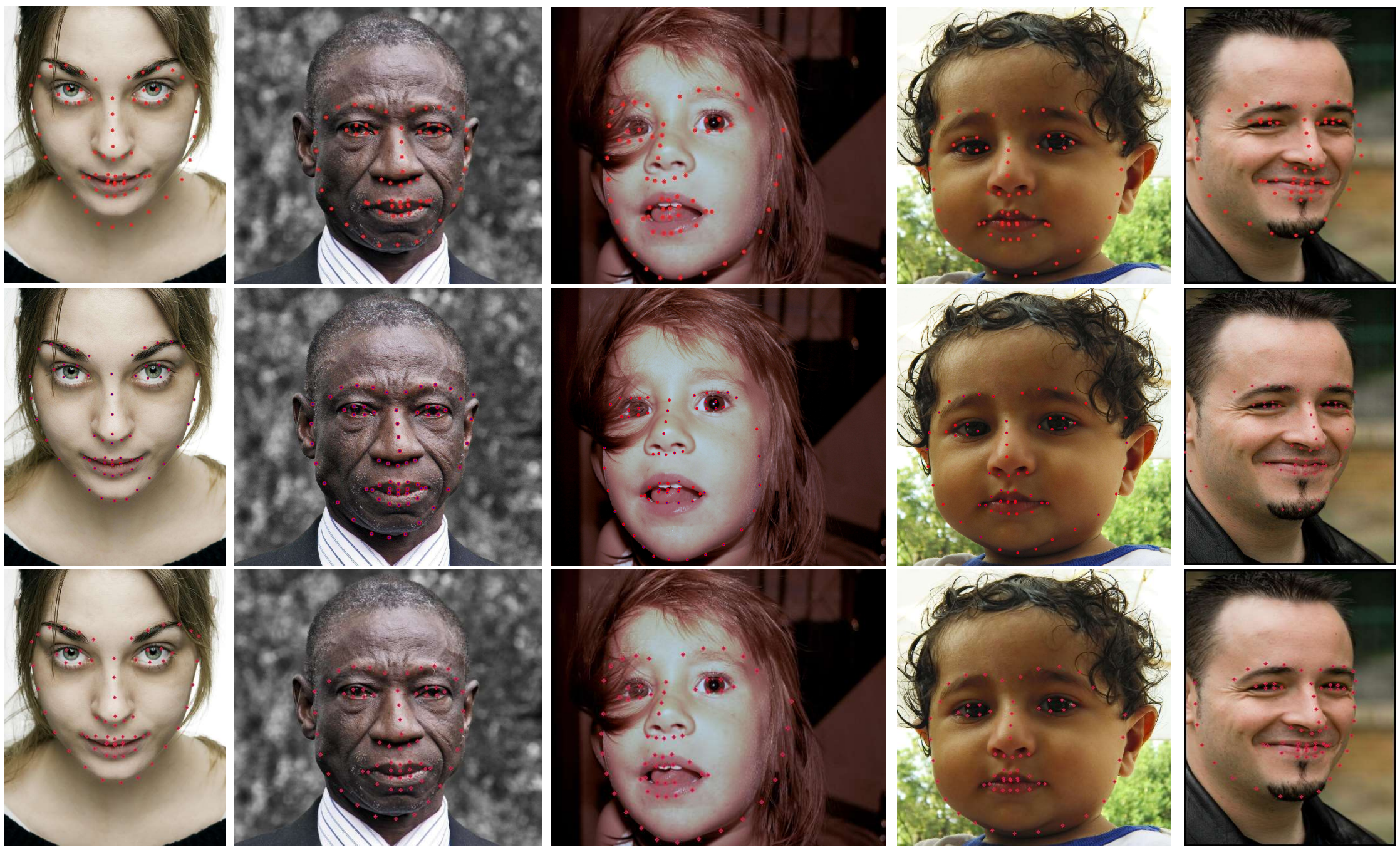

Fig. 7: Fitting results comparison (68 points, Helen dataset [21]), Row-1:LBF[1]: Row-2:CLM[18], Row-3: elBCLM

## 7. CONCLUSIONS

Two main innovations are proposed in this paper, one is that it is first time to connect two schools of face alignments into one framework, and another is PDM non-rigid local parameters are worked as discriminative features as a clue during the cascade alignment framework.